\documentclass{article}

\usepackage{microtype}
\usepackage{graphicx}
\usepackage{subfigure}
\usepackage{booktabs} 
\usepackage{comment}
\usepackage{pdfpages}

\usepackage{hyperref}

\usepackage[accepted]{icml2025}

\usepackage{amsmath}
\usepackage{amssymb}
\usepackage{mathtools}
\usepackage{amsthm}

\usepackage[T1]{fontenc} 
\usepackage[most]{tcolorbox}
\usepackage{subcaption}
\usepackage{enumitem}
\makeatletter
\gdef\@fpheader{}
\makeatother

\usepackage[capitalize,noabbrev]{cleveref}

\theoremstyle{plain}

\theoremstyle{definition}

\theoremstyle{remark}

\usepackage[textsize=tiny]{todonotes}

\begin{document}

\twocolumn[
\icmltitle{Open Source Planning \& Control System with Language Agents for Autonomous Scientific Discovery}



\icmlsetsymbol{equal}{*}
\icmlsetsymbol{supervised}{\dagger}

\begin{icmlauthorlist}
\icmlauthor{Licong Xu}{ia,kavli}
\icmlauthor{Milind Sarkar}{iiser}
\icmlauthor{Anto I.~Lonappan}{ucsd}
\icmlauthor{Íñigo Zubeldia}{ia,kavli}
\icmlauthor{Pablo Villanueva-Domingo}{cvc}
\icmlauthor{Santiago Casas}{ttk}
\icmlauthor{Christian Fidler}{ttk}
\icmlauthor{Chetana Amancharla}{infosys}
\icmlauthor{Ujjwal Tiwari}{infosys}
\icmlauthor{Adrian Bayer}{flatiron,princ}
\icmlauthor{Chadi Ait Ekioui}{camphy,paris}
\icmlauthor{Miles Cranmer}{ia,kavli,damtp}
\icmlauthor{Adrian Dimitrov}{camphy}
\icmlauthor{James Fergusson}{damtp}
\icmlauthor{Kahaan Gandhi}{camphy,haverford,caltech}
\icmlauthor{Sven Krippendorf}{damtp,camphy}
\icmlauthor{Andrew Laverick}{camphy}
\icmlauthor{Julien Lesgourgues}{ttk}
\icmlauthor{Antony Lewis}{sussex}
\icmlauthor{Thomas Meier}{lmu}
\icmlauthor{Blake Sherwin}{kavli,damtp}
\icmlauthor{Kristen Surrao}{col}
\icmlauthor{Francisco Villaescusa-Navarro}{flatiron,princ}
\icmlauthor{Chi Wang}{google}
\icmlauthor{Xueqing Xu}{camphy}
\icmlauthor{Boris Bolliet}{equal,kavli,camphy}
\icmlcorrespondingauthor{Boris Bolliet}{bb667@cam.ac.uk}

\end{icmlauthorlist}
\icmlaffiliation{google}{Google DeepMind}
\icmlaffiliation{paris}{Télécom SudParis - 9 rue Charles Fourier - 91011 Évry cedex - France}
\icmlaffiliation{haverford}{Haverford College, Haverford, PA, USA 19041}
\icmlaffiliation{caltech}{Division of Physics, Mathematics and Astronomy, California Institute of Technology, Pasadena, CA 91125, USA}
\icmlaffiliation{lmu}{MCML - Munich Center for Machine Learning, LMU Munich, Geschwister-Scholl-Platz 1, 80539 Munich, Germany}
\icmlaffiliation{sussex}{Department of Physics \& Astronomy, University of Sussex, Brighton BN1 9QH, UK}
\icmlaffiliation{ia}{Institute of Astronomy, University of Cambridge, Cambridge, United Kingdom}
\icmlaffiliation{kavli}{Kavli Institute for Cosmology, University of Cambridge, Cambridge, United Kingdom}
\icmlaffiliation{ttk}{Institute for Theoretical Particle Physics and Cosmology (TTK), RWTH Aachen University, Aachen, Germany}
\icmlaffiliation{cvc}{Computer Vision Center, Universitat Autònoma de Barcelona, Bellaterra, Barcelona, Spain}
\icmlaffiliation{iiser}{Department of Physical Sciences, Indian Institute of Science Education and Research (IISER), Mohali, Punjab, India}
\icmlaffiliation{camphy}{Department of Physics, University of Cambridge, Cambridge, United Kingdom}
\icmlaffiliation{infosys}{Infosys Ltd}
\icmlaffiliation{damtp}{Department of Applied Mathematics and Theoretical Physics (DAMTP), University of Cambridge, Cambridge, United Kingdom}
\icmlaffiliation{flatiron}{Center for Computational Astrophysics, Flatiron Institute, New York, NY, USA}
\icmlaffiliation{princ}{Department of Astrophysical Sciences, Princeton University, Princeton, NJ, USA}
\icmlaffiliation{col}{Department of Physics, Columbia University, New York, NY, USA}
\icmlaffiliation{ucsd}{Department of Physics, University of California, San Diego, CA, USA}


\icmlkeywords{Machine Learning, ICML}

\vskip 0.3in
]



\printAffiliationsAndNotice{\icmlEqualContribution} 

\begin{abstract}
We present a multi-agent system for automation of scientific research tasks, \href{https://github.com/CMBAgents/cmbagent}{\texttt{cmbagent}}. The system is formed by about 30 Large Language Model (LLM) agents and implements a \textit{Planning \& Control} strategy to orchestrate the agentic workflow, with no \textit{human-in-the-loop} at any point. Each agent specializes in a different task (performing retrieval on scientific papers and codebases, writing code, interpreting results, critiquing the output of other agents) and the system is able to execute code locally. We successfully apply \texttt{cmbagent} to carry out a PhD level cosmology task (the measurement of cosmological parameters using supernova data) and evaluate its performance on two benchmark sets, finding superior performance over state-of-the-art LLMs. The source code is available on GitHub\footnote{\href{https://github.com/CMBAgents/cmbagent}{https://github.com/CMBAgents/cmbagent}}, demonstration videos are also available\footnote{\href{https://www.youtube.com/@cmbagent}{https://www.youtube.com/@cmbagent}}, and the system is deployed on HuggingFace\footnote{\href{https://huggingface.co/spaces/astropilot-ai/cmbagent}{https://huggingface.co/spaces/astropilot-ai/cmbagent}} and will be available on the cloud\footnote{\href{https://cmbagent.cloud/}{https://cmbagent.cloud/}}.
\end{abstract}

\section{Introduction}
\label{intro}
Rapid progress in the development of Large Language Models (LLMs) is enabling new approaches to scientific research \citep[see, e.g.,][and references therein]{lu2024aiscientist, ghareeb2025robinmultiagentautomatingscientific}. This includes quantitative disciplines grounded in numerical data, such as physics, chemistry, biology, and economics, at both fundamental and applied levels. 

One promising application of LLMs in astrophysics and cosmology is their use as autonomous agents to support and automate data analysis workflows. In \citet{Laverick:2024fyh}, it was demonstrated how LLM agents can be used to solve a state-of-the-art data analysis task in cosmology, measuring the values of fundamental parameters that describe the Universe using a novel dataset \citep{ACT:2023kun}. They introduced a multi-agent system composed of several types of LLM agents specializing in different tasks (coding, retrieving information from scientific papers, using domain-specific software libraries). This multi-agent system operated with a \textit{human-in-the-loop} at every step: each LLM response was reviewed by the user, who then provided guidance for the next action. 

Although such systems can be used as effective research assistant tools, they are limited by their reliance on continuous human input. 
In this work, we introduce \texttt{cmbagent}, a multi-agent system which can carry end-to-end research tasks with no \textit{human-in-the-loop} at any point. To enable full automation, \texttt{cmbagent} employs a robotics-inspired \textit{Planning \& Control} strategy, with an agentic framework powered by \texttt{AG2} \citep{wu2023autogen, AG2_2024}. We first describe the system architecture and usage (Sec.\,\ref{sec:methods}) and then present a series of evaluations of the system demonstrating the system's effectiveness in cosmological applications and beyond (Sec.\,\ref{sec:results}), before concluding (Sec.\,\ref{sec:discussion}).


\section{Description of the system}
\label{sec:methods}

\subsection{Planning \& Control Strategy}
\label{sec:planningcontrol}
\begin{figure}[htbp]
\centering\includegraphics[width=0.5\textwidth]{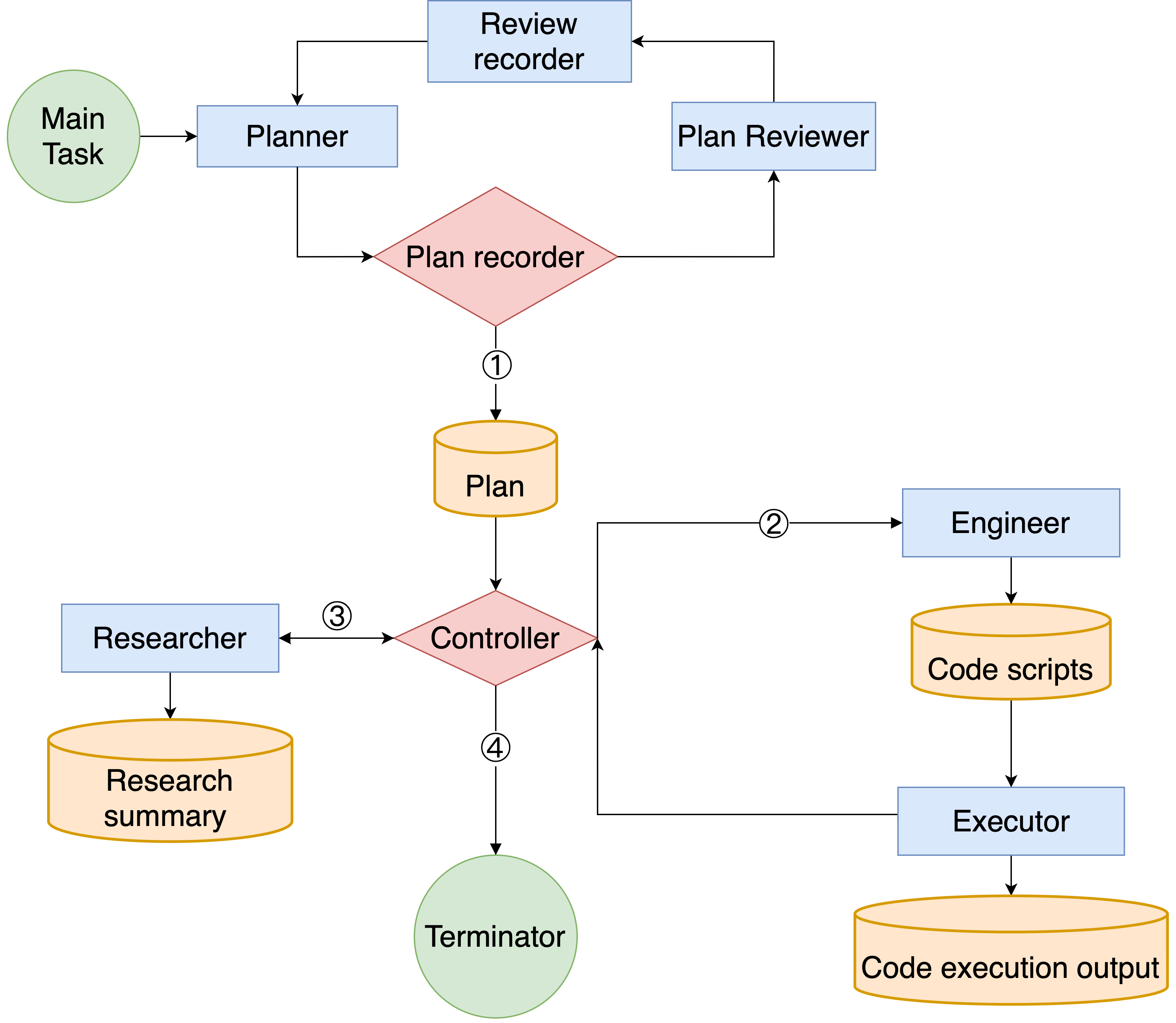}
\caption{\textit{Planning \& Control} strategy in \texttt{cmbagent}. The yellow disks correspond to output of the system. 
1: Plan approved after maximum review round is reached; 
2: Subtask involves coding and previous execution failed, but retry attempts remain; 
3: Subtask involves researching or reasoning over external knowledge; 
4: All subtasks completed or maximum code execution attempts reached.}

    \label{fig:agent_frame}
\end{figure}

The \textit{Planning \& Control} strategy 
in \texttt{cmbagent} is as follows. Given an input task, the system first goes through a \textit{Planning} phase in which a plan is designed and approved. A \textit{Control} phase follows, in which the plan is executed by the \textit{Control} agents. 
The agents involved in both phases are LLMs and there is no \textit{human-in-the-loop} at any point in either phase. The full \textit{Planning \& Control} strategy
, which is described next, is illustrated in Fig.\,\ref{fig:agent_frame}.

\paragraph{Planning phase.}
The input of the \textit{Planning} phase is the \textit{Main Task}, which is specified by the user and is typically a quantitative research task. There are two main agents in this phase: the \textit{planner}, which proposes a plan, and the \textit{plan reviewer}, which provides feedback on the proposed plan. 
 
The \textit{Planning} phase starts with the \textit{planner} proposing a first version of the plan. The plan is formed by a succession of a maximum of $n_\mathrm{steps}$ steps, with each step consisting of: (i) a sub-task, (ii) a set of actions to be taken to carry out the sub-task, and (iii) an agent in charge of completing these actions. 
The proposed plan is then reviewed by the \textit{plan reviewer}, which provides feedback on the proposed plan. This feedback is passed on to the \textit{planner}, which proposes an updated plan. This \textit{planner}--\textit{plan reviewer} loop takes place $n_\mathrm{reviews}$ times. The hyperparameters $n_\mathrm{steps}$ and $n_\mathrm{reviews}$ can be specified by the user. 



To improve performance, the output of both agents is structured by formatting agents whose sole role is to structure the text in a specific format. Also, for traceability, the structured responses of each agent (the plans and the plan feedback) are recorded into the session context by recorder agents (\textit{plan recorder} and \textit{review recorder}).

Once the definitive plan is approved by the \textit{reviewer}, it is stored into context and saved as a JSON file, and the total cost of the phase (in USD) is displayed and stored. This brings the \textit{Planning} phase to completion and \textit{Control} phase starts. 


\paragraph{Control phase.} The input of the \textit{Control} phase is the final plan approved in the \textit{Planning} phase. The key agent of the \textit{Control} phase is the \textit{controller}, which distributes the assigned sub-tasks in the plan to the relevant agents. These sub-tasks are carried out by two main agents: (i) a \textit{researcher} for reasoning, interpretation, and summarizing tasks; and (ii) an \textit{engineer} for coding tasks (by default, in Python).

When the \textit{researcher} is queried, its output is formatted in Markdown by a formatting agent and then saved by an \textit{executor} agent so the other agents can access it. On the other hand, when the \textit{engineer} is called, a \textit{nested chat} is triggered, of which only the output is kept. This nested chat is a sequence of two agents: (i) a formatting agent that structures the code written by the \textit{engineer} and provides minor adjustments (e.g., formatting issues, plot labels), and (ii) an \textit{executor} agent that executes the provided code block locally.

The execution output is then forwarded to a post-execution \textit{interpreter} agent, which decides on which agent to transfer to. A successful code execution typically triggers a transition to the \textit{control} agent, which updates the necessary context variables and proceeds with the next step of the plan. On the other hand, a failed code execution may trigger different transitions. If the number of failed execution is less than $n_\mathrm{fails}$ (a hyperparameter that can be specified by the user), the subsystem typically transits back to the \textit{engineer} for another attempt, with suggestions provided by the \textit{interpreter}. If the failure is caused by a missing Python package, the subsystem typically transits to an \textit{installer} agent that runs a bash \texttt{pip\,\,install} command before the execution is attempted again. If the number of failed executions reaches $n_\mathrm{fails}$, the system transitions to a \textit{terminator} agent and the full session ends. 

Upon termination of each sub-task, the cost is displayed and the context variables that have been updated by the agents are recorded and passed on to the start of the next step. This process is repeated over all the steps in the plan.

Throughout the \textit{Control} phase, the \textit{controller}, \textit{engineer} and \textit{researcher} are made aware of the output of the previous steps in the plan via a common block injected into their system message. The injected message contains the final code and the execution messages of the previous steps, as well as the messages produced by the \textit{researcher}. At the end of every step in the plan, the agents and the entire chat history are reset and only the system context is carried over to the next step. This strategy allows for the system to keep memory between steps while reducing the cost of the session significantly (typically, by about a factor of two) relative the case in which the agents and chat history are not reset after every step.

\subsection{Context Agents}\label{sec:camb-context}
A key element of \texttt{cmbagent} is the presence of agents with knowledge of scientific papers and domain-specific Python libraries that are relevant to the research task specified by the user. The simplest way to specialize an LLM agent to carry out a particular task is to provide contextual information relevant to the task as part of its context. One could therefore consider supplying entire papers or an entire codebase into the context of the relevant LLM-agents. This approach is limited by cost and, more fundamentally, by the LLM's input token limits. However, for the most recent models, such as \texttt{gpt-4.1} or \texttt{gemini-2.5-pro/flash}, the current input token limit stands at about a million tokens, which allows for passing hundred-page-long articles, entire codebases, or long Python package documentations into the LLM context.

Following this approach, in \texttt{cmbagent} we have built two \textit{context} agents, i.e., agents with extended context in their system message, one for each of the two most widely-used Python packages in Cosmology: \href{https://camb.readthedocs.io/en/latest/}{\texttt{camb}} 
~\citep{Lewis:1999bs} and \href{http://class-code.net/}{\texttt{class}} \citep{class}. For \texttt{class}, we manually created a Markdown document that is then passed as a single string to the \textit{class context} agent (see \href{https://github.com/santiagocasas/clapp/blob/main/classy_docs.md}{this url}). For \texttt{camb}, the relevant Markdown document is automatically generated upon each build of the package \textit{Read the Docs} documentation using \href{https://pypi.org/project/sphinx-markdown-builder/}{\texttt{sphinx\_markdown\_builder}}. The document is then collected from \href{https://camb.readthedocs.io/en/latest/_static/camb_docs_combined.md}{this URL}. This approach implies that the \textit{camb context} agent keeps updating its system message as the \texttt{camb} library evolves. 

The performance of the \textit{camb context} agent on Cosmology problems is evaluated in Sec.\,\ref{ssec:context_evals}, where we find significant enhancement relative to pre-trained models with no additional context.

\subsection{Retrieval Augmented Generation Agents}

As discussed above, providing large context to LLMs comes at both a monetary and a latency cost, with an upper limit set by the LLM maximum number of input tokens. This limit may be hit, e.g., if the system is requested to access hundreds of scientific papers or large codebase. A powerful and complementary approach to the \textit{context} agents described in Sec.\,\ref{sec:camb-context} is to build \textit{RAG} agents. These agents leverage Retrieved Augmented Generation (RAG) methods, with their contexts made of chunks of information that are retrieved from a vector embedding of the full database using similarity search (e.g., semantic or lexical search). 

Several agents in \texttt{cmbagent} adopt this strategy. These include the \textit{classy}-\textit{sz} agent, which is used in the end-to-end research example discussed in Sec.\,\ref{ssec:end-to-end}. For this agent, we created a vector embedding of the documentation of the \texttt{classy\_sz} library \citep{Bolliet:2023eob}, on which the agent performs RAG. We have also built RAG agents for databases consisting of corpora of scientific papers.


\subsection{Distribution}

In addition to the source code on GitHub, we distribute the wheels for \texttt{cmbagent} on PyPi and provide support for containerization via Docker (see \href{https://github.com/cmbagent/cmbagent?tab=readme-ov-file#docker}{our documentation}).
We have also developed a GUI for user-friendly interaction with the system, which is deployed on a dedicated \href{https://huggingface.co/spaces/astropilot-ai/cmbagent}{HuggingFace space}. For illustration, snapshots of the preliminary GUI can be found in Appendix\,\ref{app:gui_app}.

\begin{figure*}[h!]
\centering\includegraphics[width=0.9\textwidth]{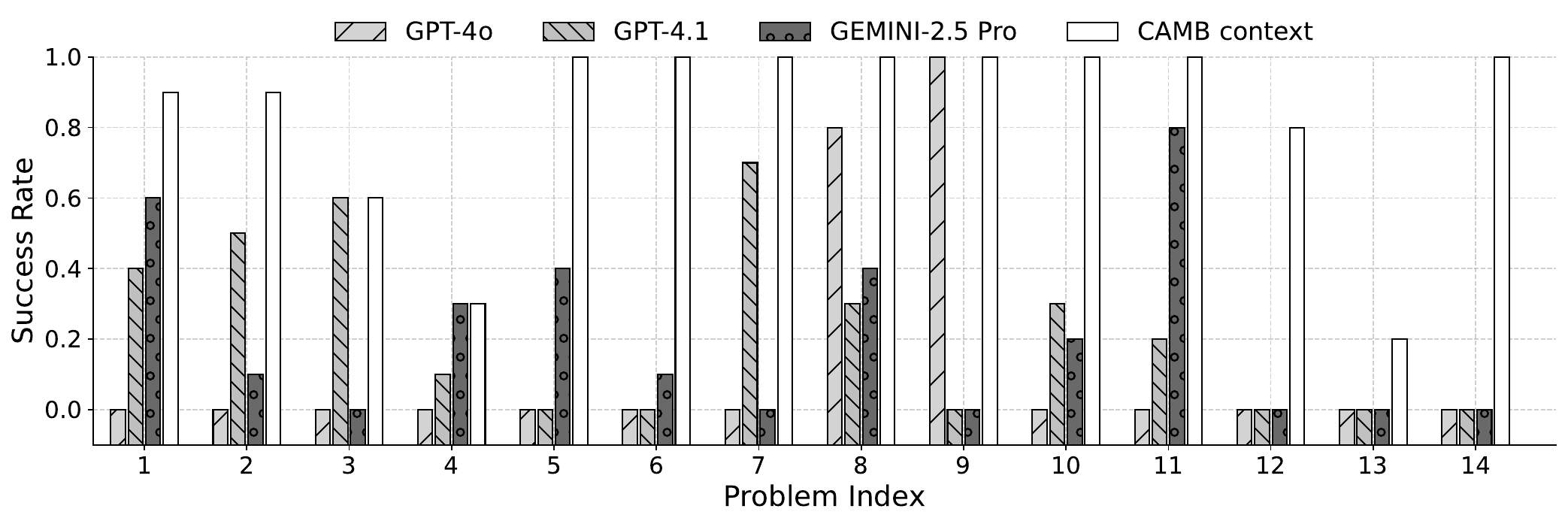}
    \setlength{\abovecaptionskip}{-2pt}\caption{\label{fig:eval_camb_context} Success rate of the \textit{camb context} agent on 14 cosmology problems, compared with state-of-the-art LLMs queried via the \textit{engineer} agent. The code used for evaluation and the full problem set are available  \href{https://github.com/cmbagent/Benchmarks}{here}. 
    }
    \label{fig:camb_context}
\end{figure*}

\section{System evaluation}
\label{sec:results}

\subsection{Context agent evaluation}
\label{ssec:context_evals}

We evaluate \texttt{cmbagent}'s \textit{camb agent} (running with \texttt{gemini-2.5-pro}) with a benchmark set formed by 14 problems involving the \texttt{camb} library, one of the most commonly-used libraries for Cosmology-specific computations. These problems range from basic calculations, such as Cosmic Microwave Background (CMB) power spectra estimation, to advanced problems, such as computing CMB delensing efficiency. We compare the performance of our \textit{camb agent} to that of three LLMs queried through the \textit{engineer} agent: \texttt{gpt-4o}, \texttt{gpt-4.1} and \texttt{gemini-2.5-pro}. We compute the success rate for each problem by solving it ten times with each agent. The results are shown on Fig.~\ref{fig:camb_context}, where it can be seen that the \textit{camb agent} compares very favorably to the other three ones. In particular, in Problems 12, 13, and 14, \texttt{gpt-4o}, \texttt{gpt-4.1} and \texttt{gemini-2.5-pro} fail systematically but the \textit{camb context} agent achieves much better performance. These results clearly demonstrate the effectiveness of domain-specific context augmentation.

\subsection{Evaluation of \textit{One Shot} and \textit{Planning \& Control} on the DS-1000 Benchmark}

We evaluate \texttt{cmbagent} on a subset of problems from the DS-1000 benchmark \citep{lai2022ds1000naturalreliablebenchmark}, which covers usage of the \texttt{pandas}, \texttt{numpy}, and \texttt{matplotlib} libraries, with the \textit{researcher} agent powered by \texttt{gpt-4.1}. 
The results are reported in Table \ref{tab:ds1000-eval}, where it can be seen that our \textit{Planning \& Control} strategy consistently leads to enhanced performance, increasing the overall success rate (number of problems solved over number of problems) from 66\% to 78\%. These results demonstrate the power of our \textit{Planning \& Control} strategy for complex problem solving. See Appendix~\ref{app:ds-1000} for further details.

\subsection{Cosmology research task}
\label{ssec:cosmo}
We also demonstrate the performance of \texttt{cmbagent} on a Cosmology research task using the \textit{Planning \& Control} mode, with default settings and \texttt{gpt-4.1} as the LLM backend. 
The task is to perform cosmological parameter estimation using the Union2.1 Type Ia supernovae dataset \citep{suzuki2012hubble}. 
This task requires conducting parameter posterior inference via Markov chain Monte Carlo (MCMC) sampling, which is the most widely used technique in Cosmology for fitting theoretical models to observational data. 

The task was successfully solved the first time it was run. Details of the output are provided in Appendix \ref{sec:appendix_sne}, including the full \textit{Main Task}, and the generated plan and plots (Figs.~\ref{fig:sne_hubble} and ~\ref{fig:sne_post}). We recorded the full session, with an accelerated version of the video available on YouTube.\footnote{\href{https://www.youtube.com/watch?v=oooLqnfB8ds}{YouTube Video: Cosmology Research Task.}} The full log is also available in a Jupyter notebook.\footnote{\href{https://github.com/cmbagent/cmbagent/blob/main/docs/notebooks/cmbagent_beta3_supernovae.ipynb}{Jupyter Notebook: Cosmology Research Task.}} 



\subsection{End-to-end research backend}
\label{ssec:end-to-end}
\texttt{cmbagent} is integrated in the soon-to-be-released\footnote{As of July 9th 2025 the package is not publicly released.} \texttt{denario} project \citep{AP_2025}, a multi-agent system designed for conducting autonomous end-to-end scientific research. There, \texttt{cmbagent} plays the role of the research results generation backend. In \texttt{denario}, a problem or data description is provided by the user. The system then follows a \textit{Planning \& Control} strategy to generate ideas through a conversation between an \textit{idea maker} and an \textit{idea hater} agent, which ends with a proposed research idea. A methodology is then generated (by two rounds of a \textit{researcher} agent); next, the proposed research project is carried out; and, finally, the results are collected. The output consists of markdown reports and plots. These are in turn, converted into a publication-ready research manuscript in PDF format by another multi-agent subsystem implemented with \href{https://www.langchain.com/langgraph}{\texttt{LangGraph}}. Literature search and populating the manuscript with relevant references are also done automatically using a \href{ttps://docs.perplexity.ai/home}{Perplexity} agent with \texttt{sonar-reasoning-pro}. 
The results page of the preliminary \texttt{denario} GUI is shown in Appendix~\ref{app:end_to_end} and an example of a  generated paper is appended to this manuscript (title: \textit{Regime-Specific Performance of 1D CNN and FCNN Architectures for Non-linear Matter Power Spectrum Emulation in $\Lambda$CDM Cosmology}).

\section{Discussion}
\label{sec:discussion}

We have introduced \texttt{cmbagent}, an LLM-powered multi-agent system for quantitative research. For complex tasks, \texttt{cmbagent} follows a \textit{Planning \& Control} strategy that allows the system to achieve full automation with no \textit{human-in-the-loop}, also offering offers \textit{One Shot} and \textit{Human-in-the-loop} modes for less complex tasks. Key features of \texttt{cmbagent} include agents specialising in research papers and code libraries, feedback loops between pairs of agents, structured output generation, and the ability to execute code locally. Built upon the \texttt{AG2} framework, its agents leverage state-of-the-art LLMs through API interfaces with OpenAI, Google, and Anthropic. The source code is available on GitHub, distributed on PyPi, and deployed on HuggingFace.


The performance results presented in Sec.\,\ref{sec:results} are remarkable. Achieving results such as the cosmological parameter estimation task discussed in Sec.\,\ref{ssec:cosmo} was hardly possible with a \textit{human-in-the-loop} and frontier LLMs only a year ago \citep[e.g.,][]{Bassett2024Integrals},\footnote{We thank David I. Kaiser for bringing this to our attention.} let alone fully autonomously. As we have demonstrated, current LLMs, when orchestrated in multi-agent systems such as ours, can solve such tasks with no difficulty. We are now even able to write research-quality papers automatically \citep[e.g.,][]{lu2024aiscientist,Moss2025,AP_2025}. These promising results and the rapid progress in AI suggest that scientific research is likely to undergo drastic changes in the near future, with a significant part of research workflows likely to be automated. This may come at a high cost \citep[e.g.][]{9869565}, and as scientists we must be lucid and address the ethical and environmental challenges associated with these changes.

\section*{Impact Statement}

We present a multi-agent system designed to carry out quantitative scientific research tasks, offering potential benefits in scalability, rigor, and reproducibility. However, such systems raise concerns about automation bias, fairness, and the potential for misuse or misinterpretation of findings. To address these risks, we emphasize the importance of transparency, robust evaluation, and oversight. Ensuring responsible use will require careful consideration of how these systems influence scientific workflows and trust in their outputs.

\section*{Author Contributions}

The development of \texttt{cmbagent} was led by BB. The development of \texttt{denario} (Section \ref{ssec:end-to-end}) was equally led by FVN, BB, PVD and will be presented in detail in a separate manuscript. LX led the development of the GUI, based on previous work by SC, CF, FVN. IZ, LX and BB wrote the paper. MS, AIL, XX and AD developed the  benchmark datasets and evaluations. PVD, CA, UT led the deployment. AB developed the idea agents. KS developed the cost report interface and led early developments of the \textit{swarm} orchestration. KG developed the integration of vision language models and CAE the context-based agents, with AL (Lewis), SC and CF. AL, AL, BS, CW, JF, JL, MC, SK, TM provided crucial input at various stages of the project.

\section*{Acknowledgments}
This work was partially funded by an unrestricted gift from Google, the Cambridge Centre for Data-Driven Discovery Accelerate Programme and the Infosys-Cambridge AI Centre. We are very grateful to the referees and panel of the ICML 2025 ML4ASTRO workshop for reviewing and accepting our work. We are very grateful to the AG2 community.\footnote{\href{https://ag2.ai/}{https://ag2.ai}}

\bibliographystyle{icml2025}
\bibliography{reference}

\appendix

\clearpage
\section{Graphical User Interface}
\label{app:gui_app}

\begin{figure*}[h!]
\centering
    \includegraphics[width=1\textwidth]{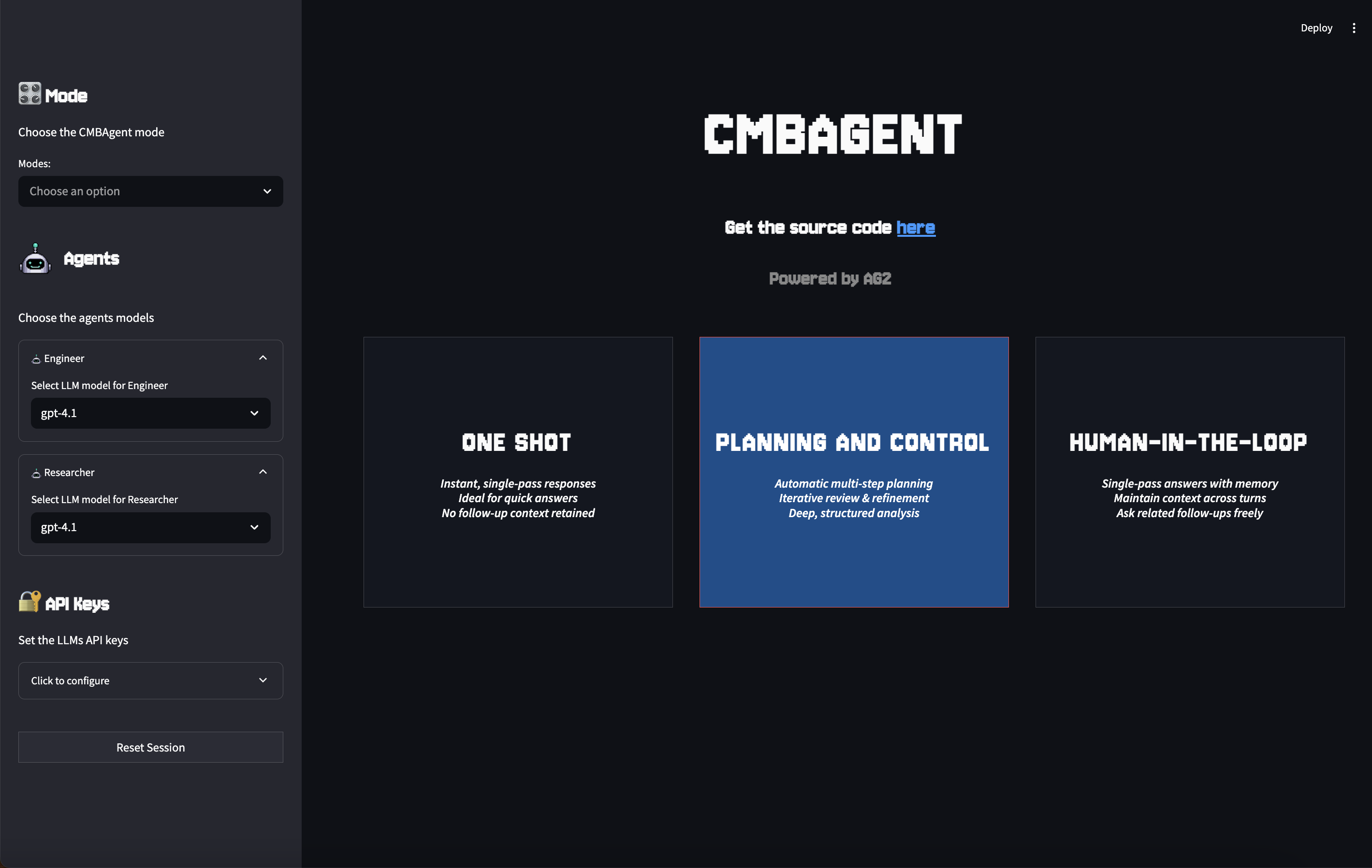}
    \caption{\label{fig:gui} Welcome page of the \texttt{cmbagent} GUI.
    }
\end{figure*}

\begin{figure*}[h!]
\centering
    \includegraphics[width=1\textwidth]{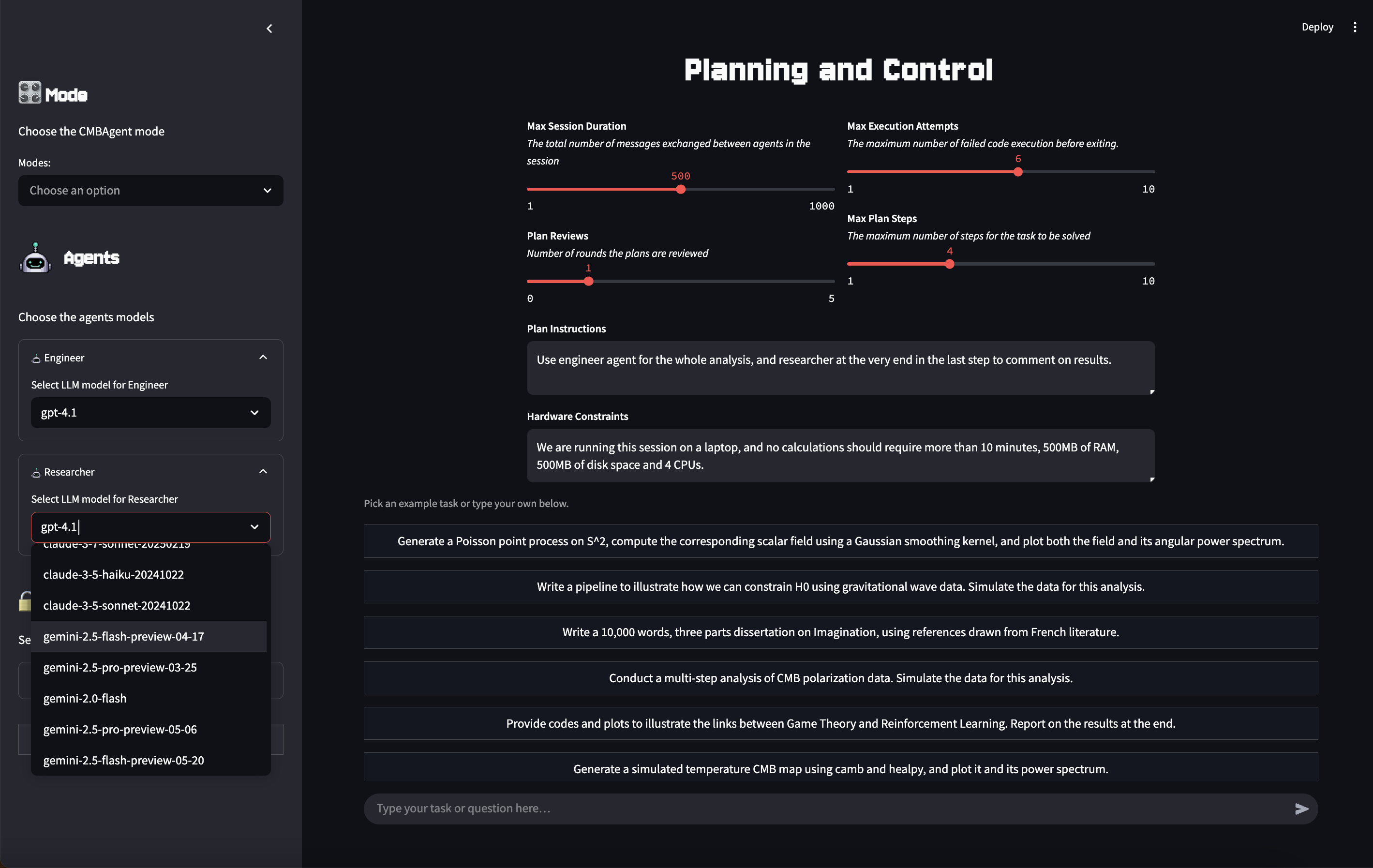}
    \caption{\label{fig:pc} \textit{Planning \& Control} page of the \texttt{cmbagent} GUI.
    }
\end{figure*}

\begin{figure*}[h!]
\centering
    \includegraphics[width=1\textwidth]{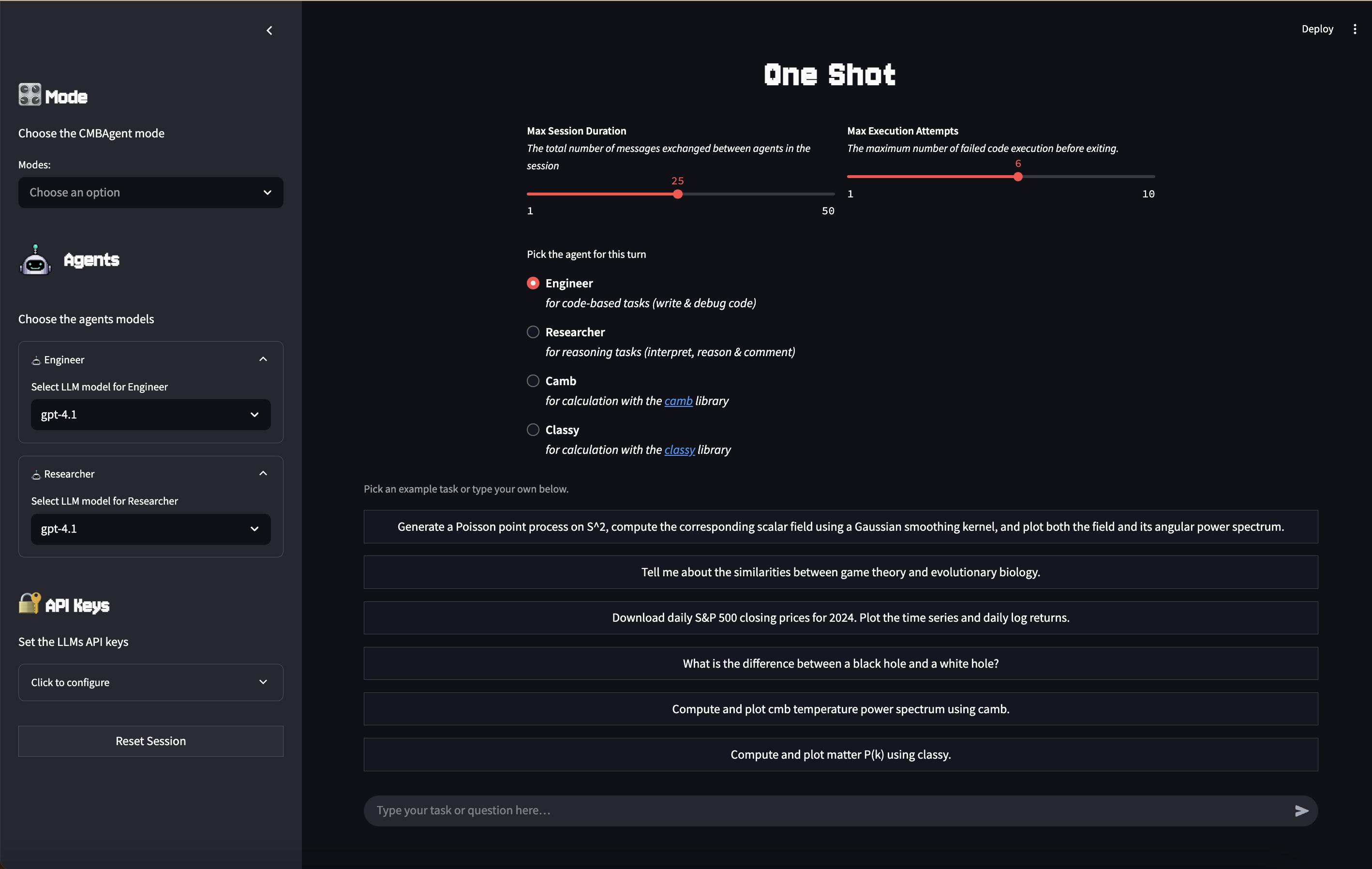}
    \caption{\label{fig:os} \textit{One Shot} page of the \texttt{cmbagent} GUI.
    }
\end{figure*}

\begin{figure*}[h!]
\centering
    \includegraphics[width=1\textwidth]{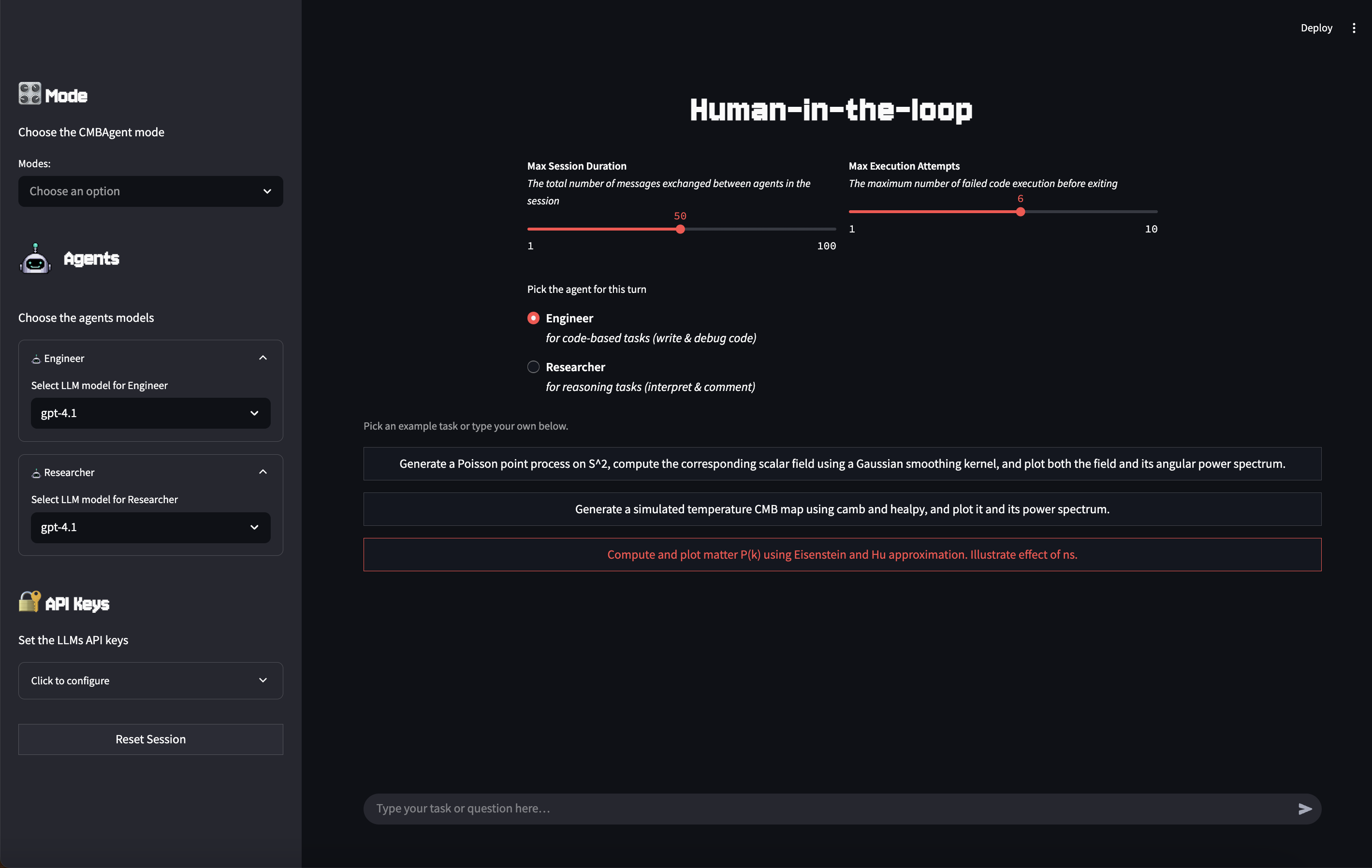}
    \caption{\label{fig:hitl} \textit{Human-in-the-loop} page of the \texttt{cmbagent} GUI.
    }
\end{figure*}

\texttt{cmbagent}'s GUI welcome page is shown in Fig.~\ref{fig:gui}. The \textit{Planning \& Control} page is shown in Fig.~\ref{fig:pc}, the \textit{One Shot} one in Fig.~\ref{fig:os}, and the \textit{Human-in-the-loop} one in Fig.~\ref{fig:hitl}.

\clearpage
\section{DS-1000 Benchmark Evaluation Table}
\label{app:ds-1000}

\begin{table}[ht]
\centering
\setlength{\arrayrulewidth}{0.7pt}
\renewcommand{\arraystretch}{1.1}
\begin{tabular}{|l|c|c|c|}
\hline
\textbf{Library} & \textbf{Problems} & \textbf{One Shot} & \textbf{Planning \& Control} \\
\hline
pandas & 0--10 & 0.3 & 0.6 \\
pandas & 10--20 & 0.8 & 0.8 \\
pandas & 30--40 & 0.8 & 0.9 \\
numpy & 300--310 & 0.9 & 0.9 \\
matplotlib &600--610 & 0.5 & 0.7 \\
\hline
\textbf{Total} & -- & \textbf{0.66} & \textbf{0.78} \\
\hline
\end{tabular}
\caption{Comparison of \texttt{cmbagent} One Shot and Planning \& Control strategies on subsets of DS-1000 \citep{lai2022ds1000naturalreliablebenchmark}. The evaluation notebook is available \href{https://github.com/cmbagent/Benchmarks/blob/main/DS1000evalCMBAgent_Planning_and_control.ipynb}{here}.}
\label{tab:ds1000-eval}
\end{table}

\clearpage
\section{Cosmology example: Analysis of Union2.1 Type Ia supernovae data}
\label{sec:appendix_sne}
\begin{figure*}[h!]
\centering
    \includegraphics[width=0.7\textwidth]{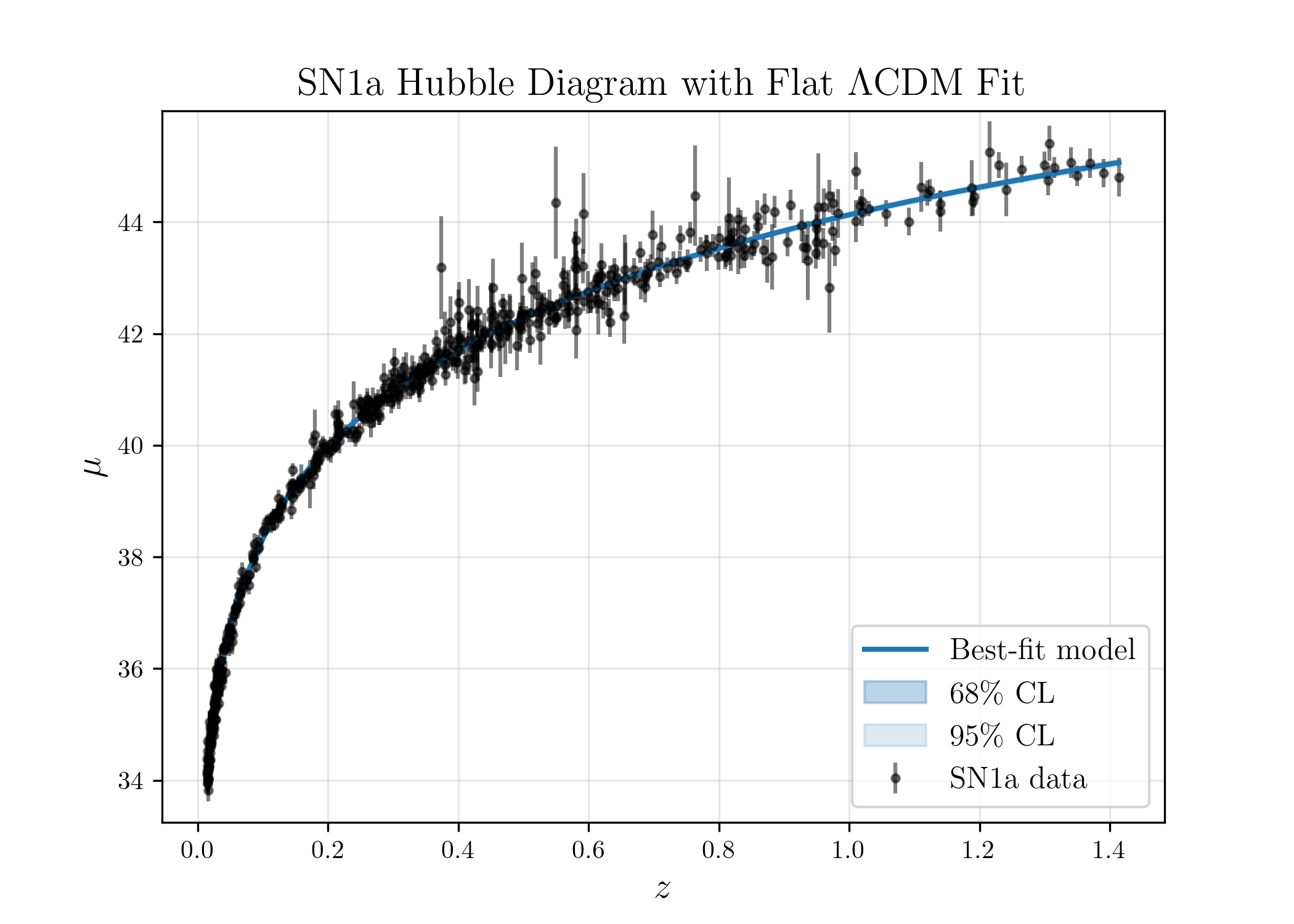}
    \caption{\label{fig:sne_hubble} Distance modulus versus redshift for 580 SNe Ia for the Union2.1 supernova sample, and the best fitted curve to the data.}
\end{figure*}

\begin{figure*}[h!]
\centering
    \includegraphics[width=0.5\textwidth]{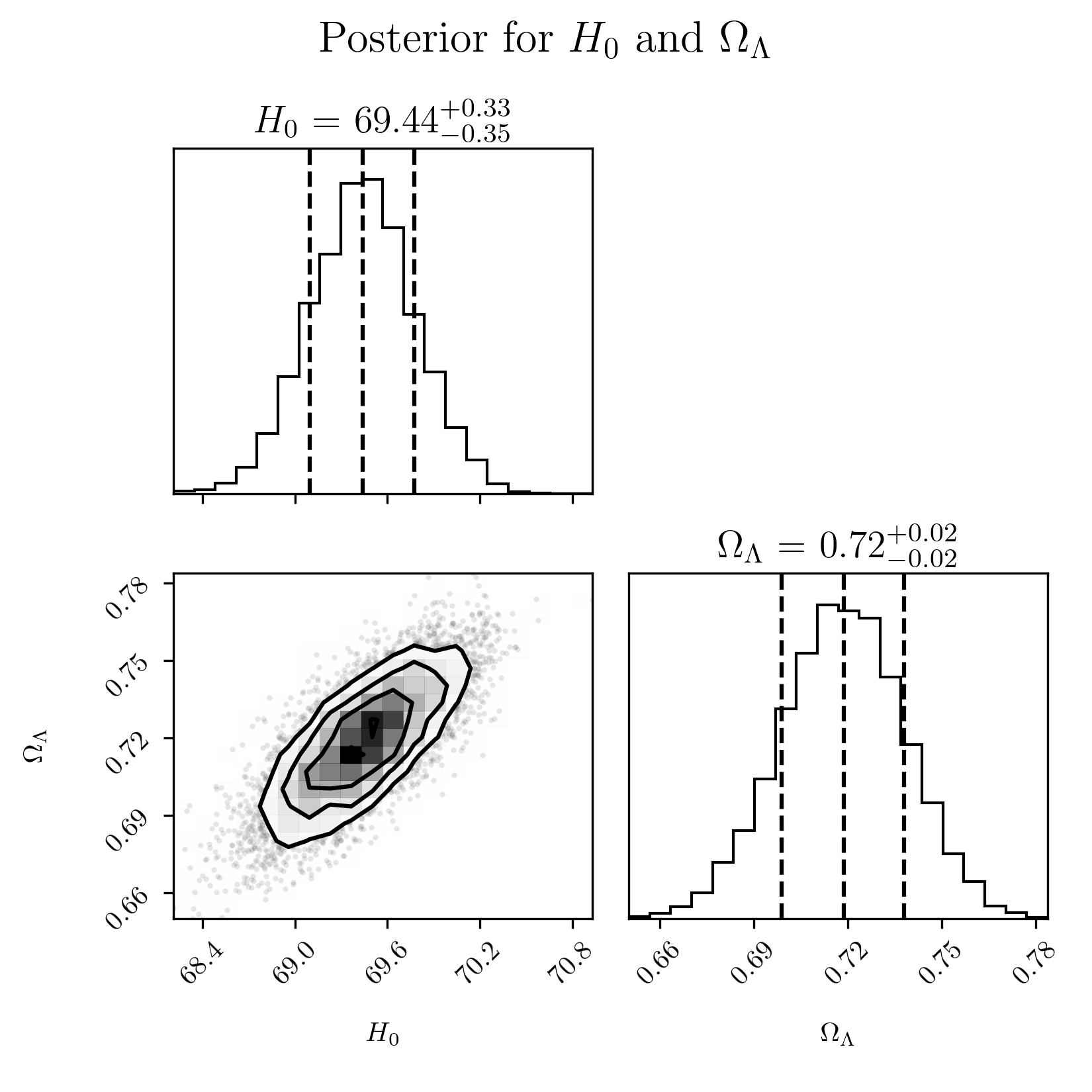}
    \caption{\label{fig:sne_post} Two-dimensional marginalized constraints on $H_0$ and $\Omega_\Lambda$, derived from the Union2.1 supernova sample.}
\end{figure*}

In this Appendix we show the \textit{Main Task} and \texttt{cmbagent}'s output in the \textit{Planning} phase for the Cosmology task discussed in Sec.~\ref{ssec:cosmo}. In the \textit{Control} phase, \texttt{cmbagent} produced a plot of the distance modulus versus redshift for the Union2.1 SNe sample, which is shown in Fig. \ref{fig:sne_hubble}. The posteriors for parameters $H_0$ and $\Omega_\Lambda$ are also shown in Fig. \ref{fig:sne_post}.

\begin{tcolorbox}[colback=blue!5!white, colframe=blue!75!black,title=Main Task, sharp corners=south]
Download the file:\\
https://supernova.lbl.gov/Union/figures/SCPUnion\\2.1\_mu\_vs\_z.txt\\
Its description is:\\
<description>\\
An ASCII table with tab-separated columns: Supernova Name, Redshift, Distance Modulus, and Distance Modulus Error. For Union2.1, there is an additional column for the probability that the supernova was hosted by a low-mass galaxy.\\
</description>\\
Fit this data within flat LCDM model with two free parameters: H0 and Omega\_L. Write a simple MCMC (but optimized/fast) code to fit for H0 and Omega\_L using the SN1a data.\\
Make a contour plot and show the 1d posteriors, and quote the mean and 1-sigma on each parameter. 
Finally show the data along with the best fit model and 68\%/95\% CL regions. 
Comment on the results.\\ 
\textbf{Constraints:} We are running this analysis on a Macbook Pro with 8 available threads. Ensure you use the resources optimally so the MCMC can run fast, i.e., within a few minutes until convergence.
Have the engineer agent do a preliminary MCMC timing step in a separate step. 
\end{tcolorbox}

\begin{tcolorbox}[colback=green!5!white, colframe=green!75!black, title=Final Plan, sharp corners=southwest, breakable]
\textbf{Step 1: Download and preprocess the SN1a data}
\begin{itemize}[itemsep=2pt, parsep=2pt, topsep=2pt]
    \item \textbf{Sub-task:} Download and preprocess the SN1a data
    \item \textbf{Agent:} Engineer
    \item \textbf{Instructions:}
    \begin{itemize}[itemsep=2pt, parsep=2pt, topsep=2pt]
        \item Download the \texttt{SCPUnion2.1\_mu\_vs\_z.txt} file from the provided URL.
        \item Parse the ASCII table, extracting Supernova Name, Redshift, Distance Modulus, Distance Modulus Error, and the fifth column (probability of low-mass galaxy host).
        \item Verify the data format, especially the fifth column, and determine whether the probability of low-mass galaxy host should be incorporated into the analysis (e.g., as a weight or for a subsample).
        \item Store the data in a suitable format (e.g., numpy arrays or pandas DataFrame) for analysis.
        \item Perform basic data quality checks (e.g., remove NaNs, check for outliers).
    \end{itemize}
\end{itemize}

\textbf{Step 2: Implement the flat LCDM model, likelihood function, and priors}
\begin{itemize}[itemsep=2pt, parsep=2pt, topsep=2pt]
    \item \textbf{Sub-task:} Implement the flat LCDM model, likelihood function, and priors
    \item \textbf{Agent:} Engineer
    \item \textbf{Instructions:}
    \begin{itemize}[itemsep=2pt, parsep=2pt, topsep=2pt]
        \item Write a function to compute the theoretical distance modulus for given $H_0$ and $\Omega_L$ in a flat LCDM cosmology.
        \item Implement the likelihood function comparing model predictions to observed data, accounting for errors.
        \item Define and justify appropriate priors for $H_0$ and $\Omega_L$ to be used in the MCMC analysis.
    \end{itemize}
\end{itemize}

\textbf{Step 3: Preliminary MCMC timing and optimization}
\begin{itemize}[itemsep=2pt, parsep=2pt, topsep=2pt]
    \item \textbf{Sub-task:} Preliminary MCMC timing and optimization
    \item \textbf{Agent:} Engineer
    \item \textbf{Instructions:}
    \begin{itemize}[itemsep=2pt, parsep=2pt, topsep=2pt]
        \item Set up a simple MCMC sampler for $H_0$ and $\Omega_L$.
        \item Benchmark different MCMC implementations (e.g., \texttt{emcee}, \texttt{PyMC}, custom Metropolis-Hastings) to determine which performs best with 8 available threads.
        \item Run a short test chain using the selected implementation to estimate runtime and convergence speed.
        \item Adjust sampler settings (number of walkers, steps, parallelization) for optimal performance on a Macbook Pro.
        \item Report estimated time to convergence.
    \end{itemize}
\end{itemize}

\textbf{Step 4: Run the full MCMC analysis}
\begin{itemize}[itemsep=2pt, parsep=2pt, topsep=2pt]
    \item \textbf{Sub-task:} Run the full MCMC analysis
    \item \textbf{Agent:} Engineer
    \item \textbf{Instructions:}
    \begin{itemize}[itemsep=2pt, parsep=2pt, topsep=2pt]
        \item Run the optimized MCMC sampler to obtain posterior samples for $H_0$ and $\Omega_L$.
        \item Implement burn-in removal and thinning of chains if necessary to ensure independent samples.
        \item Ensure convergence diagnostics are met (e.g., check trace plots, autocorrelation).
        \item Save the posterior samples for further analysis.
    \end{itemize}
\end{itemize}

\textbf{Step 5: Generate plots and summary statistics}
\begin{itemize}[itemsep=2pt, parsep=2pt, topsep=2pt]
    \item \textbf{Sub-task:} Generate plots and summary statistics
    \item \textbf{Agent:} Engineer
    \item \textbf{Instructions:}
    \begin{itemize}[itemsep=2pt, parsep=2pt, topsep=2pt]
        \item Create a corner plot showing the 2D posterior contours and 1D marginalized posteriors for $H_0$ and $\Omega_L$.
        \item Compute and report the mean and 1-sigma uncertainties for each parameter.
        \item Plot the SN1a data with the best-fit model and 68\%/95\% confidence regions.
        \item Calculate the reduced chi-square or other goodness-of-fit metrics to quantify the quality of the best-fit model.
    \end{itemize}
\end{itemize}

\textbf{Step 6: Comment on the results}
\begin{itemize}[itemsep=2pt, parsep=2pt, topsep=2pt]
    \item \textbf{Sub-task:} Comment on the results
    \item \textbf{Agent:} Researcher
    \item \textbf{Instructions:}
    \begin{itemize}[itemsep=2pt, parsep=2pt, topsep=2pt]
        \item Interpret the best-fit values and uncertainties for $H_0$ and $\Omega_L$.
        \item Discuss the quality of the fit and the implications for cosmology.
        \item Comment on the consistency with other measurements and any limitations of the analysis.
        \item Specifically discuss the tension (if any) between the derived $H_0$ value and recent measurements from other methods like the Planck CMB and local distance ladder.
    \end{itemize}
\end{itemize}
\end{tcolorbox}

\clearpage
\section{RAG Prompts}
\label{app:rag_prompts}

Our modified SciRagPaperQA2 prompt priorities conciseness and domain specificity for efficient human evaluation.
\begin{tcolorbox}[colback=blue!5!white, colframe=blue!75!black, title={\textbf{SciRagPaperQA2 Prompt}}]
Provide a concise answer in 1-2 sentences maximum. 

Context (with relevance scores):\{context\}

Question: \{question\}

Write a concise answer based on the context, focusing on astronomical facts and concepts. 
If the context provides insufficient information, reply

\{CANNOT\_ANSWER\_PHRASE\}.

Write in the style of a scientific astronomy reference, with precise and 
factual statements. The context comes from a variety of sources and is 
only a summary, so there may be inaccuracies or ambiguities. 

\{prior\_answer\_prompt\} Answer (maximum one sentence):
\end{tcolorbox}

\noindent In contrast, the original prompt emphasizes comprehensive information synthesis, mandatory citation and Wikipedia-style formatting.
\begin{tcolorbox}[colback=blue!5!white, colframe=blue!75!black, title={\textbf{Original PaperQA2 Prompt}}]
Answer the question below with the context.

Context (with relevance scores):{context}

Question: \{question\}

Write an answer based on the context. 
If the context provides insufficient information reply \{CANNOT\_ANSWER\_PHRASE\}

For each part of your answer, indicate which sources most support 
it via citation keys at the end of sentences, like \{example\_citation\}.

Only cite from the context above and only use the citation keys from the context. 
\{CITATION\_KEY\_CONSTRAINTS\}

Do not concatenate citation keys, just use them as is. 

Write in the style of a Wikipedia article, with concise sentences and 
coherent paragraphs. The context comes from a variety of sources and is 
only a summary, so there may inaccuracies or ambiguities. If quotes are 
present and relevant, use them in the answer. This answer will go directly 
onto Wikipedia, so do not add any extraneous information.

\{prior\_answer\_prompt\}Answer (\{answer\_length\}):
\end{tcolorbox}
The SciRagHybrid adopts a structured approach, requiring a JSON format return for consistent response parsing.
\begin{tcolorbox}[colback=blue!5!white, colframe=blue!75!black, title={\textbf{SciRagHybrid System Prompt}}]
You are a helpful assistant. Answer based on the provided context. 
You must respond in valid JSON format with the following structure:

\{
  "answer": "your detailed answer here",
  "sources": ["source1", "source2", "source3"]\}

The sources must be from the **Context** material provided. 
Include source names, page numbers, equation numbers, table 
numbers, section numbers when available. Ensure your response 
is valid JSON only.
\end{tcolorbox}
Finally, the SciRagOpenAI system uses a tool-based retrieval approach with markdown formatting, emphasising precise source and knowledge integration.
\begin{tcolorbox}[colback=blue!5!white, colframe=blue!75!black, title={\textbf{SciRagOpenAI System Prompt}}]
You are a retrieval agent. 
You must add precise source from where you got the answer.
Your answer should be in markdown format with the following 
structure: 

**Answer**:\{answer\}

**Sources**:\{sources\}

You must search your knowledge base calling your tool. The 
sources must be from the retrieval only.
You must report the source names in the sources field, if 
possible, the page number, equation number, table number, 
section number, etc.
\end{tcolorbox}
These distinct prompts demonstrate different strategies for balancing response quality, source attributions, and output formatting in scientific RAG systems.

\clearpage
\section{End-to-End Research}
\label{app:end_to_end}
\begin{figure*}[h!]
\centering
    \includegraphics[width=1\textwidth,trim=0cm 2cm 0cm 0cm, clip]{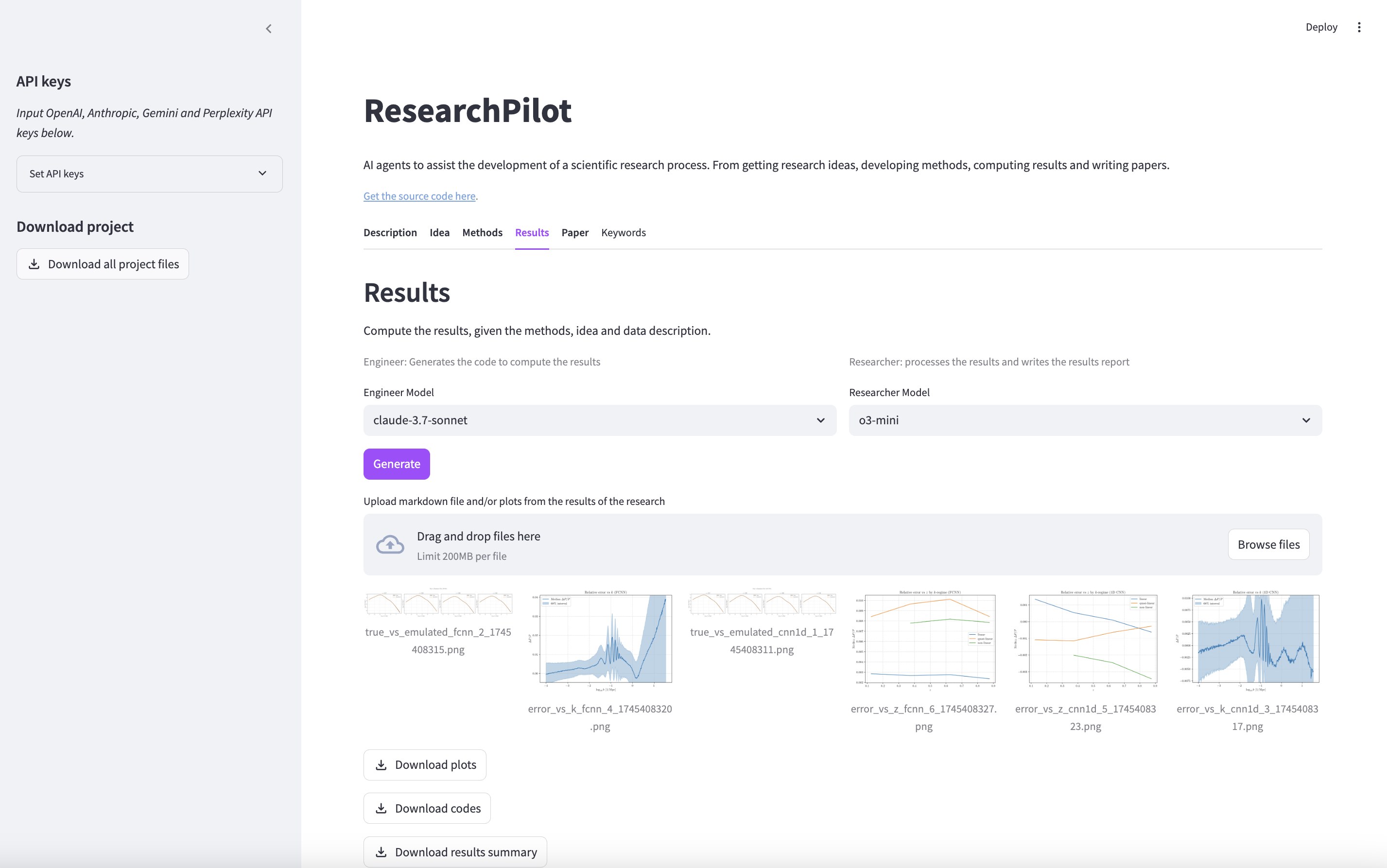}
    \caption{The (preliminary) results page of the \texttt{denario} GUI.}
    \label{fig:gui_ap}
\end{figure*}

Fig.~\ref{fig:gui_ap} shows the results page of the \texttt{denario} GUI. The example paper, fully automatically generated, is appended.

\section*{Supplementary material}

# Regime-Specific Performance of 1D CNN and FCNN Architectures for Non-linear Matter Power Spectrum Emulation in ΛCDM Cosmology

AstroPilot[1]

[1]*Anthropic, Gemini & OpenAI servers. Planet Earth.*

## ABSTRACT

Accurate and efficient estimation of the non-linear matter power spectrum is essential for maximizing cosmological constraints from upcoming large-scale structure surveys. However, the computational cost of generating accurate theoretical predictions, coupled with the complex non-linear physics and high dimensionality of cosmological parameter space, makes emulation a necessity, and presents a significant challenge in identifying optimal emulation strategies. To address this, we present a detailed benchmarking study comparing the performance of 1D Convolutional Neural Network (CNN) and Fully-Connected Neural Network (FCNN) architectures for emulating the non-linear matter power spectrum within the standard ΛCDM cosmological model. Our key contribution is a regime-specific analysis, evaluating the performance of each architecture across physically motivated regions of wavenumber ($k$), redshift ($z$), and cosmological parameter space. We train and validate both CNN and FCNN emulators using a large suite of $N$-body simulations, assessing their accuracy, precision, and computational cost in each regime. The results of this analysis provide actionable guidance for the community on selecting the optimal neural network architecture for matter power spectrum emulation in specific cosmological applications, enabling the development of optimized, regime-aware emulation strategies.

*Keywords:* Cosmology, Cosmological parameters

## 1. INTRODUCTION

Accurate and efficient estimation of the non-linear matter power spectrum is crucial for extracting maximal cosmological information from upcoming large-scale structure surveys, such as the Vera C. Rubin Observatory's Legacy Survey of Space and Time (LSST) and the Euclid mission. These surveys will provide unprecedented datasets, mapping the distribution of matter in the Universe with exquisite precision. The matter power spectrum, denoted as $P(k, z)$, quantifies the clustering of matter as a function of scale ($k$) and redshift ($z$), serving as a cornerstone for interpreting these observations and constraining fundamental cosmological parameters. Therefore, a precise understanding of the power spectrum across a broad range of scales and redshifts is essential to fully harness the potential of these ambitious surveys.

Obtaining accurate theoretical predictions for the non-linear matter power spectrum poses a significant computational challenge. Traditional methods, such as $N$-body simulations, are computationally expensive, requiring substantial resources and time to generate a single power spectrum for a given set of cosmological parameters (Silva et al. 2024,?). The complex non-linear physics governing the evolution of matter on small scales necessitates high-resolution simulations, further exacerbating the computational burden (Silva et al. 2024,?). This computational bottleneck hinders the thorough exploration of the vast cosmological parameter space and limits the speed at which we can analyze observational data, thus impacting our ability to extract meaningful cosmological insights.

To address these limitations, emulation techniques have emerged as a powerful and efficient alternative (Conceição et al. 2023; Jense et al. 2024). Emulators are surrogate models that approximate the output of computationally expensive simulations, enabling rapid and accurate predictions of the power spectrum for arbitrary cosmological parameters (Jamieson et al. 2024; Günther et al. 2025). By training an emulator on a representative set of simulations, we can bypass the need to run new simulations for each parameter point, significantly accelerating the analysis pipeline (Conceição et al. 2023; Jamieson et al. 2024). This acceleration is particularly crucial for Markov Chain Monte Carlo (MCMC) analyses, which require evaluating the power spectrum for millions of parameter combinations to accurately sam-

ple the posterior distribution of cosmological parameters (Jamieson et al. 2024; Günther et al. 2025).

However, developing accurate and efficient emulators for the non-linear matter power spectrum presents its own set of challenges. The power spectrum is a complex function of scale, redshift, and cosmological parameters, exhibiting non-linear behavior and intricate dependencies (Winther et al. 2019; Sáez-Casares et al. 2023). Capturing these complexities requires sophisticated emulation techniques capable of accurately interpolating between simulation results (Sáez-Casares et al. 2023; Fremstad & Winther 2025). Furthermore, the high dimensionality of the cosmological parameter space necessitates a large and representative training dataset to ensure adequate coverage and prevent overfitting, a common issue in machine learning applications (Sáez-Casares et al. 2023; Fremstad & Winther 2025).

Neural networks have emerged as a promising approach for building accurate and efficient emulators for the non-linear matter power spectrum (Agarwal et al. 2012; Trusov et al. 2025). Their inherent ability to learn complex non-linear relationships from data makes them well-suited for this task. Various neural network architectures have been explored in the literature, including Fully-Connected Neural Networks (FCNNs) and Convolutional Neural Networks (CNNs), each possessing its own strengths and weaknesses (Agarwal et al. 2012; Trusov et al. 2025). Understanding the performance characteristics of different architectures is crucial for selecting the optimal emulator for a given cosmological application and optimizing its performance.

In this work, we present a detailed benchmarking study comparing the performance of 1D CNN and FCNN architectures for emulating the non-linear matter power spectrum within the standard $\Lambda$CDM cosmological model (Bourilkov 2019; Saxena et al. 2024). Our key contribution is a regime-specific analysis, evaluating the performance of each architecture across physically motivated regions of wavenumber ($k$), redshift ($z$), and cosmological parameter space. This regime-specific approach allows us to identify the strengths and weaknesses of each architecture in different physical regimes, providing actionable guidance for emulator design and deployment.

Specifically, we train and validate both CNN and FCNN emulators using a large suite of matter power spectrum data generated using the 'classy_sz' code, a modified version of the Cosmic Linear Anisotropy Solving System (CLASS) code. Our training data is meticulously sampled using Latin Hypercube Sampling (LHS) to ensure uniform coverage of the cosmological parameter space, mitigating potential biases and improving the generalizability of the emulators. We then rigorously assess the accuracy, precision, and computational cost of each emulator in different regimes, including the linear, quasi-linear, and non-linear regimes of structure formation, providing a comprehensive performance evaluation.

Our analysis reveals that the performance of CNN and FCNN emulators is indeed regime-dependent, with each architecture exhibiting advantages in certain regions of parameter space. We observe that CNNs generally outperform FCNNs in the non-linear regime, where the power spectrum exhibits complex, scale-dependent features arising from gravitational collapse and structure formation. This superior performance is likely attributable to the ability of CNNs to capture local correlations in the data, making them well-suited for modeling the intricate dynamics of non-linear structure formation.

Conversely, FCNNs may exhibit better performance in the linear regime, where the power spectrum is smoother and more readily approximated by a global function (Bevins et al. 2025; Facchinetti 2025). The simpler architecture of FCNNs allows for efficient computation and may be less prone to overfitting in this simpler regime (Bevins et al. 2025). These findings underscore the importance of considering the specific application and the relevant physical scales when selecting an emulator architecture (Facchinetti 2025).

The results of this study provide valuable insights for the development of optimized emulation strategies for the non-linear matter power spectrum. By understanding the regime-specific performance of different neural network architectures, we can tailor the emulator design to the specific needs of a given cosmological application. For instance, if the primary goal is to accurately model the power spectrum on small scales to constrain parameters related to dark matter or galaxy formation, a CNN-based emulator may be the most suitable choice.

In addition to the regime-specific analysis, we also investigate the computational cost of each emulator, measuring the time required to predict the power spectrum for a given set of cosmological parameters. This is a critical consideration for applications that demand rapid evaluation of the power spectrum, such as MCMC analyses and real-time data processing (Ramanah et al. 2020; Yang et al. 2025). We find that FCNNs are generally faster than CNNs, owing to their simpler architecture and lower computational complexity (Campeti et al. 2025). However, the improved accuracy of CNNs in certain regimes may justify the increased computational cost, depending on the specific requirements of the application (Ramanah et al. 2020).

Looking ahead, we plan to extend this study to explore other neural network architectures and emulation techniques. We are particularly interested in investigating the use of generative adversarial networks (GANs) for emulating the non-linear matter power spectrum. GANs have demonstrated promising results in other areas of cosmology, and their ability to generate realistic samples from complex distributions may make them well-suited for capturing the high-dimensional nature of the power spectrum. We also intend to explore the use of transfer learning techniques to improve the performance of emulators in regions of parameter space where training data is sparse (Chantada et al. 2023; Gómez-Vargas et al. 2023). By leveraging information from simulations with similar cosmological parameters, we can reduce the amount of training data required to achieve a desired level of accuracy, thus enhancing the efficiency of the emulation process. Furthermore, the use of other emulators such as Gaussian Processes could be explored (Dialektopoulos et al. 2022).

## 2. METHODS

### 2.1. *Data Generation*

#### 2.1.1. *Cosmological Parameter and Redshift Sampling*

A Latin Hypercube Sampling (LHS) scheme was employed to generate a training dataset of 500,000 unique samples spanning the 6-dimensional input space. The cosmological parameters, specifically the baryon density ($\omega_b$), cold dark matter density ($\omega_{cdm}$), Hubble constant ($H_0$), amplitude of the primordial power spectrum ($\log A$), and spectral index ($n_s$), along with the redshift ($z$), were varied within the following ranges: $\omega_b \in [0.01933, 0.02533]$, $\omega_{cdm} \in [0.08, 0.20]$, $H_0 \in [40, 100]$ km/s/Mpc, $\log A \in [2.5, 3.5]$, $n_s \in [0.8, 1.2]$, and $z \in [0, 1]$. Each sample was represented as a vector of the form ($\omega_b, \omega_{cdm}, H_0, \log A, n_s, z$). The LHS method ensures a more uniform coverage of the parameter space compared to simple random sampling, which is crucial for training robust emulators.

#### 2.1.2. *Power Spectrum Computation*

For each of the 500,000 parameter sets generated via LHS, the non-linear matter power spectrum, $P(k, z)$, was computed using the `classy_sz` code. `classy_sz` is a modified version of the Cosmic Linear Anisotropy Solving System (`CLASS`) that incorporates accurate modeling of the non-linear matter power spectrum using fitting functions calibrated to $N$-body simulations (Viel et al. 2011; Silva et al. 2024,?). The power spectra were evaluated over a fixed $k$-grid provided by `classy_sz`, spanning the range $k \in [10^{-4}, 10] \, h/\text{Mpc}$. This range encompasses both the linear and non-linear regimes of

structure formation (Viel et al. 2011). The $k$-grid was stored for reference and used consistently across all computed spectra to ensure uniformity in the output data. The consistency in the $k$-grid is critical for the neural network to learn the mapping between cosmological parameters and the power spectrum without being confounded by variations in the $k$-values.

#### 2.1.3. *Dataset Structuring and Splitting*

The generated dataset consists of paired input and output data. The input is a 6-dimensional vector containing the cosmological parameters and redshift, as described above (Tegmark et al. 2006). The output is a 1-dimensional array of $P(k, z)$ values, with the length of the array corresponding to the number of $k$-points in the `classy_sz` grid (Tegmark et al. 2006; Oddo et al. 2021). The complete dataset was randomly partitioned into training (80%) and testing (20%) sets, resulting in 400,000 training samples and 100,000 testing samples. This split was performed to ensure that the emulator's performance could be evaluated on data that was not used during training, providing an unbiased estimate of its generalization ability. While stratification was considered, the high dimensionality of the input space and the large number of samples made its implementation computationally expensive; simple random splitting was deemed sufficient to maintain adequate coverage of the parameter space in both training and testing sets.

### 2.2. *Data Preprocessing*

#### 2.2.1. *Input Normalization*

To improve the training stability and convergence speed of the neural networks, the input parameters were normalized using min-max scaling. Each parameter, $\omega_b$, $\omega_{cdm}$, $H_0$, $\log A$, $n_s$, and $z$, was scaled to the range $[0, 1]$ according to the formula:

$$x_{norm} = \frac{x - x_{min}}{x_{max} - x_{min}}$$

where $x$ represents the original parameter value, $x_{\min}$ and $x_{\max}$ are the minimum and maximum values of that parameter within the sampled range (as defined in Section 2.1.1), and $x_{\text{norm}}$ is the normalized value. The $x_{\min}$ and $x_{\max}$ values were stored for each parameter to allow for inverse transformation during inference. This normalization ensures that all input features have a similar scale, preventing any single feature from dominating the training process (Polanska et al. 2024).

#### 2.2.2. *Output Transformation and Normalization*

The output spectra, $P(k, z)$, were transformed to log-log space to better represent the wide range of values and

to linearize the relationship between the input parameters and the output. Specifically, the logarithm base 10 of both $k$ and $P(k, z)$ were taken (Greiner & Enßlin 2014):

$$\log_{10} P(k, z) \quad \text{as a function of} \quad \log_{10} k$$

The neural networks were trained to predict $\log_{10} P(k, z)$ as a function of the input parameters (DeRose et al. 2021; Choudhury et al. 2024; Trusov et al. 2025). Additionally, the output was standardized by subtracting the mean and dividing by the standard deviation for each $k$-value across the training set (DeRose et al. 2021; Cagliari et al. 2025). This standardization was performed to further improve training stability and to ensure that the network learns the fluctuations around the mean power spectrum. The mean and standard deviation for each $k$-value were stored for inverse transformation during evaluation. The inverse transformation is given by: (DeRose et al. 2021; Trusov et al. 2025)

$$P(k, z) = 10^{(\sigma_k \times P_{predicted} + \mu_k)}$$

where $\sigma_k$ and $\mu_k$ are the stored standard deviation and mean for the given $k$ respectively, and $P_{predicted}$ is the output of the neural network.

### 2.2.3. *Data Storage*

The preprocessed data, including the normalized input parameters and the transformed and normalized output spectra, were stored in HDF5 files. Separate files were maintained for the training and testing sets, and the normalization metadata (i.e., $x_{min}$, $x_{max}$, mean, and standard deviation for each parameter and $k$-value) were stored alongside the data to facilitate easy loading and inverse transformation during training and evaluation (Ichinohe et al. 2018; Li et al. 2024). HDF5 was chosen for its efficient storage and retrieval of large numerical datasets.

## 2.3. *Neural Network Training*

### 2.3.1. *1D CNN Architecture*

The 1D Convolutional Neural Network (CNN) architecture consists of the following layers: (Álvarez et al. 2023)

1. **Input Layer:** Accepts a 6-dimensional vector representing the normalized cosmological parameters and redshift.

2. **Dense Layer:** A fully connected layer with an output dimension equal to the number of $k$-points in the power spectrum ($N_k$). This layer uses a ReLU activation function.

3. **Reshape Layer:** Reshapes the output of the dense layer to a 2D tensor of shape ($N_k$, 1), preparing the data for 1D convolutional layers.

4. **Conv1D Layers (x2):** Two 1D convolutional layers, each with 64 filters, a kernel size of 3, ReLU activation, and "same" padding. "Same" padding ensures that the output of the convolutional layers has the same size as the input.

5. **Flatten Layer:** Flattens the output of the convolutional layers into a 1D vector.

6. **Dense Layer:** A fully connected layer with 256 units and ReLU activation.

7. **Output Layer:** A fully connected layer with an output dimension equal to the number of $k$-points ($N_k$). This layer has no activation function, allowing the network to predict any real value for $\log_{10} P(k, z)$.

### 2.3.2. *FCNN Architecture*

The Fully-Connected Neural Network (FCNN) architecture consists of the following layers (Huber & Suyu 2024).

1. **Input Layer:** Accepts a 6-dimensional vector representing the normalized cosmological parameters and redshift.

2. **Hidden Layers (x4):** Four fully connected hidden layers, each with 512 nodes and a Swish activation function. The Swish activation function is defined as $f(x) = x \cdot \text{sigmoid}(x)$.

3. **Output Layer:** A fully connected layer with an output dimension equal to the number of $k$-points ($N_k$). This layer has no activation function.

### 2.3.3. *Training Procedure*

Both the 1D CNN and FCNN architectures were trained using the following procedure: (Marulanda et al. 2020)

1. **Optimizer:** The Adam optimizer was used to minimize the loss function.

2. **Learning Rate:** An initial learning rate of $10^{-3}$ was used. The learning rate was tuned as needed during preliminary experiments.

3. **Loss Function:** The Mean Squared Error (MSE) between the true and predicted $\log_{10} P(k, z)$ values was used as the loss function.

4. **Batch Size:** A batch size of 128 was used. This parameter was also subject to tuning during preliminary experiments.

5. **Early Stopping:** Early stopping was implemented to prevent overfitting. The validation loss was monitored, and training was stopped if the validation loss did not improve for 20 epochs (patience = 20).

6. **Maximum Epochs:** A maximum of 500 epochs was allowed for training.

7. **Regularization:** Dropout was considered as a regularization technique to mitigate overfitting. If overfitting was observed, dropout layers with a rate of 0.1-0.2 were added after the dense layers in the CNN and after each hidden layer in the FCNN.

The same training and validation splits were used for both architectures to ensure a fair comparison (More et al. 2024). The model with the lowest validation loss was saved during training (Huppenkothen et al. 2023). Training curves (loss vs. epoch) were logged for later analysis (More et al. 2024).

## 2.4. *Performance Evaluation*
### 2.4.1. *Accuracy Metrics*

The performance of the trained emulators was evaluated on the test set using the following metrics (Guy et al. 2022; Huppenkothen et al. 2023; Malz et al. 2023).

1. **Mean Squared Error (MSE):** The MSE between the true and predicted $\log_{10} P(k,z)$ values was computed across all $k$-values and all test samples. This provides an overall measure of the emulator's accuracy.

2. **Relative Error:** The relative error, $\Delta P/P$, was computed as a function of $k$ and $z$, averaged over the test set. The relative error is defined as:

$$\frac{\Delta P(k,z)}{P(k,z)} = \frac{|P_{true}(k,z) - P_{predicted}(k,z)|}{P_{true}(k,z)}$$

This metric provides a more detailed assessment of the emulator's performance in different regions of $k$ and $z$ (Jones et al. 2024; Günther et al. 2025).

The error analysis was stratified by $k$-range (e.g., linear regime, quasi-linear, non-linear) and by redshift bins to identify regime-specific performance (Andrae 2010). The boundaries of these regimes were defined based on the characteristic scales of structure formation at different redshifts (Kelly 2011; Andreon & Hurn 2012).

### 2.4.2. *Inference Speed*

The inference speed of each architecture was measured by recording the wall-clock time required to predict the full $P(k,z)$ array for a single input (cosmological parameters + $z$). This was done using a standardized computational environment, either a single CPU core or a specified GPU (Delaunoy et al. 2020; Wang et al. 2023). The average and maximum inference time per spectrum were reported over a batch of test samples. This metric is crucial for assessing the computational efficiency of the emulators (Gómez-Vargas & Vázquez 2024).

### 2.4.3. *Diagnostic and Comparison Plots*

Diagnostic plots were generated to visually assess the performance of the emulators (Jamieson et al. 2024). These plots included:

1. **True vs. Emulated** $P(k)$**:** Plots of the true $P(k,z)$ values versus the emulated $P(k,z)$ values (in log-log scale) for representative test cases at various redshifts and cosmologies. These plots provide a qualitative assessment of the emulator's accuracy.

2. **Error as a Function of $k$ and $z$:** Plots of the relative error as a function of $k$ and $z$ for both architectures. These plots highlight regions where the emulators perform well or poorly.

### 2.4.4. *Reporting*

The results of the performance evaluation were summarized in tables and figures comparing the accuracy and speed of the 1D CNN and FCNN architectures across different regimes. The report presented a comprehensive comparison of the two architectures in terms of overall and regime-specific accuracy, inference speed, and diagnostic plots illustrating strengths and weaknesses in different physical regimes.

## 3. RESULTS
### 3.1. *Results*

This section presents a detailed performance analysis of the 1D CNN and FCNN architectures for emulating the non-linear matter power spectrum, $P(k,z)$, within the $\Lambda$CDM cosmological framework. The emulators were trained and validated using a large dataset of 500,000 spectra generated from a Latin Hypercube sampling of cosmological parameters and redshifts ($z \in [0,1]$), with spectra computed using the `classy_sz` Boltzmann solver. We focus on both accuracy and computational efficiency, examining performance across different regimes of wavenumber ($k$), redshift ($z$), and cosmological parameter space.

### 3.1.1. *Quantitative Performance Metrics*

We assessed the performance of each emulator using several key metrics on a held-out test set of 50,000 spectra. These metrics include the Mean Squared Error (MSE) and the median absolute relative error ($|\Delta P/P|$), calculated across various $k$ regimes and redshift bins.

*Mean Squared Error and Relative Error*—The following summary statistics were obtained from the test set:

- **1D CNN:**

  - Overall MSE (mean over all $k$): $6.08 \times 10^{-4}$
  - Linear regime ($k < 0.1\,h/\mathrm{Mpc}$):
    * MSE: $5.15 \times 10^{-5}$
    * Median $|\Delta P/P|$: 0.0037
  - Quasi-linear regime ($0.1 \leq k < 0.5\,h/\mathrm{Mpc}$):
    * MSE: $1.95 \times 10^{-4}$
    * Median $|\Delta P/P|$: 0.0076
  - Non-linear regime ($k \geq 0.5\,h/\mathrm{Mpc}$):
    * MSE: $1.59 \times 10^{-3}$
    * Median $|\Delta P/P|$: N/A (See below)

- **FCNN:**

  - Overall MSE (mean over all $k$): $8.40 \times 10^{-4}$
  - Linear regime ($k < 0.1\,h/\mathrm{Mpc}$):
    * MSE: $3.45 \times 10^{-5}$
    * Median $|\Delta P/P|$: 0.0041
  - Quasi-linear regime ($0.1 \leq k < 0.5\,h/\mathrm{Mpc}$):
    * MSE: $1.04 \times 10^{-4}$
    * Median $|\Delta P/P|$: 0.0111
  - Non-linear regime ($k \geq 0.5\,h/\mathrm{Mpc}$):
    * MSE: $2.31 \times 10^{-3}$
    * Median $|\Delta P/P|$: N/A (See below)

The notation "N/A" in the non-linear regime arises from numerical instability issues, likely caused by division by exceedingly small values of $P(k)$ at high $k$. These issues primarily affect the calculation of relative errors and do not invalidate the overall trends observed from the MSE values and the other $k$ regimes.

*Redshift-Binned Performance*—The median absolute relative error, $|\Delta P/P|$, was also calculated within specific redshift bins:

| Redshift Bin | 1D CNN | FCNN |
|---|---|---|
| [0.00, 0.25) | N/A | N/A |
| [0.25, 0.50) | 0.0052 | 0.0069 |
| [0.50, 0.75) | 0.0052 | 0.0071 |
| [0.75, 1.00) | 0.0057 | 0.0067 |

Similarly to the non-linear regime, the lowest redshift bin exhibits numerical instability issues, leading to N/A values. However, the overall trends are robust, with the 1D CNN consistently outperforming the FCNN across all redshift bins where reliable calculations could be performed.

### 3.1.2. *Inference Speed*

The inference speed of each emulator was measured on an NVIDIA A100 GPU:

- **1D CNN:** Average inference time per spectrum: 0.073 s (max: 0.357 s)

- **FCNN:** Average inference time per spectrum: 0.074 s (max: 0.314 s)

Both architectures exhibit comparable inference speeds, with sub-0.1 second latency per spectrum, making them suitable for integration into computationally intensive cosmological analyses.

### 3.1.3. *Regime-Specific Analysis*

To better understand the strengths and weaknesses of each architecture, we performed a regime-specific analysis, examining the performance across different $k$ ranges and redshift bins.

*Linear Regime ($k < 0.1\,h/\mathrm{Mpc}$)*—In the linear regime, both emulators achieve high accuracy, with median relative errors below 0.5%. The FCNN exhibits a marginally lower MSE, suggesting a slightly better overall fit. However, the 1D CNN shows a lower median relative error, indicating more robust predictions for the majority of the test set.

*Quasi-linear and Non-linear Regimes ($k \geq 0.1\,h/\mathrm{Mpc}$)*—The 1D CNN demonstrates a clear advantage in the quasi-linear and non-linear regimes. Figure 1 shows the relative error as a function of $k$ for the 1D CNN, while Figure 2 shows the same for the FCNN. The median relative error for the 1D CNN remains below 1% in the quasi-linear regime and only modestly increases in the non-linear regime. In contrast, the FCNN's error grows more rapidly with increasing $k$. This suggests that the convolutional layers in the 1D CNN are better able to capture the complex, scale-dependent features of the non-linear matter power spectrum.

*Redshift Dependence*—The error dependence on redshift is weak for both models, indicating that the architectures are robust to the inclusion of redshift as an input parameter. Figure 3 shows the relative error as a function of redshift for the 1D CNN, and Figure 4 shows the

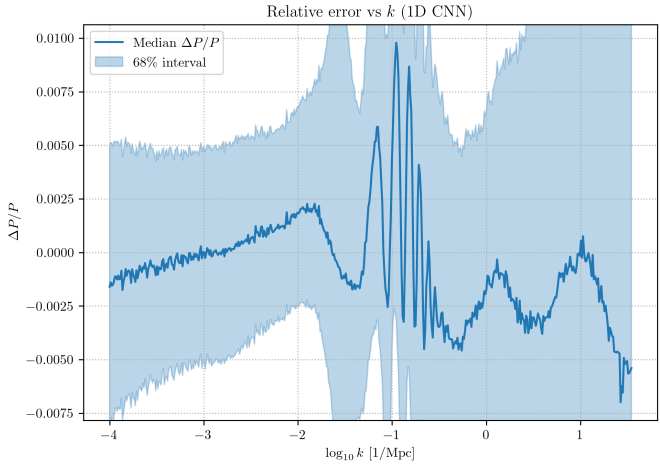

**Figure 1.** The figure shows the relative error $\Delta P/P$ as a function of $k$ for a 1D CNN. The median $\Delta P/P$ is shown as a blue line, and the 68% confidence interval is shown as a shaded blue region. Large differences are seen at around $\log_{10} k \approx -1$.

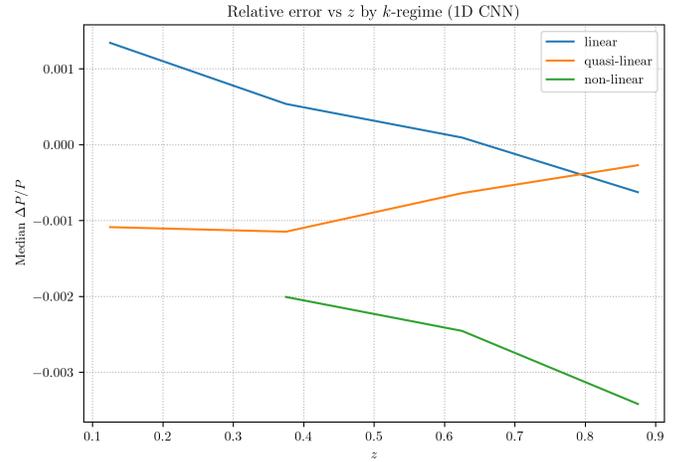

**Figure 3.** The figure shows the relative error in the power spectrum, $\Delta P/P$, as a function of redshift $z$ for different $k$-regimes (linear, quasi-linear, and non-linear) using a 1D CNN. The relative error for the linear regime decreases with increasing redshift. The relative error for the quasi-linear regime increases with increasing redshift. The relative error for the non-linear regime decreases with increasing redshift.

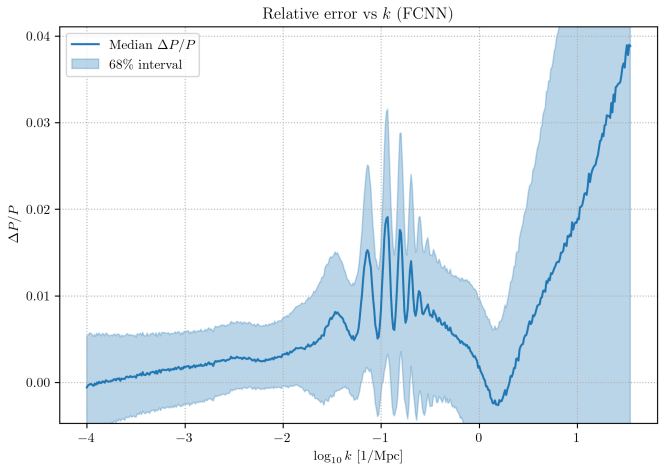

**Figure 2.** Relative error of the power spectrum $\Delta P/P$ as a function of wavenumber $k$ using a Fully Convolutional Neural Network (FCNN). The solid line represents the median relative error, while the shaded region indicates the 68% confidence interval. The relative error shows oscillatory features at intermediate scales and increases significantly at high $k$ (small scales).

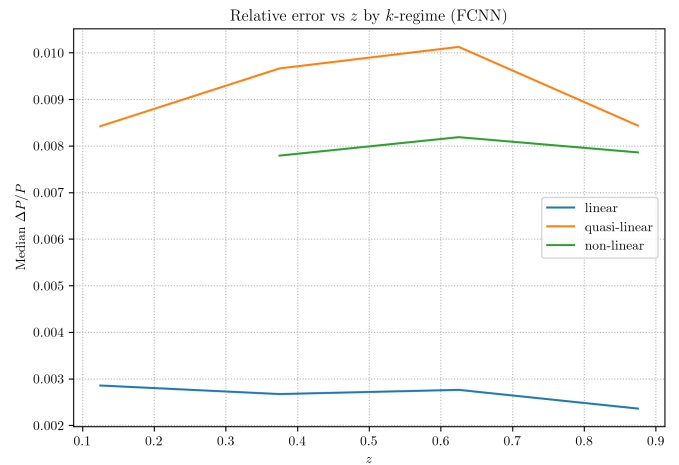

**Figure 4.** The figure shows the median relative error $\Delta P/P$ as a function of redshift $z$ for different $k$-regimes: linear, quasi-linear, and non-linear. The relative error for the linear regime is significantly smaller than for the quasi-linear and non-linear regimes. The quasi-linear regime shows the largest relative error, peaking around $z \approx 0.6$.

same for the FCNN. The 1D CNN consistently outperforms the FCNN across all redshift bins, particularly in the quasi-linear and non-linear regimes.

Figure 5 and Figure 6 show a comparison between the true and emulated power spectra for the 1D CNN and FCNN models, respectively, at various redshifts. Both models show good agreement with the true power spectrum.

### 3.1.4. *Discussion*

The results of this benchmarking study demonstrate that 1D CNN architectures provide a robust, accurate, and efficient solution for emulating the non-linear matter power spectrum in $\Lambda$CDM cosmology. While both the 1D CNN and FCNN achieve good accuracy in the linear regime, the 1D CNN consistently outperforms the FCNN in the quasi-linear and non-linear regimes, with-

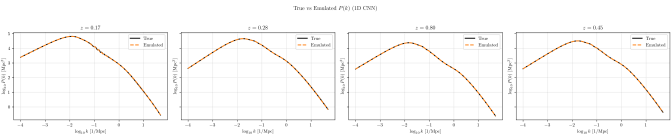

**Figure 5.** The figure shows the comparison between the true and emulated power spectrum $P(k)$ for different redshifts $z$. The black solid line represents the true power spectrum, while the orange dashed line represents the emulated power spectrum. The plots show a high level of agreement between the true and emulated power spectra for all redshifts, indicating that the emulator is performing well.

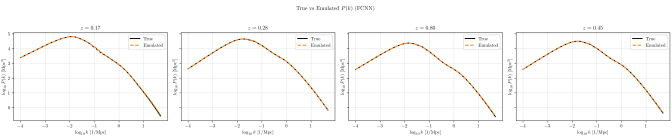

**Figure 6.** Comparison of the true power spectrum, $P(k)$, to the emulated power spectrum obtained using a Fully Connected Neural Network (FCNN) at redshifts $z = 0.17, 0.28, 0.80$, and $0.45$. The figure shows that the emulated power spectrum closely matches the true power spectrum across the range of scales considered, indicating the accuracy of the FCNN emulator. Small differences are found.

out incurring additional computational cost. This suggests that the convolutional layers in the 1D CNN allow it to better capture the complex, scale-dependent features of the non-linear matter power spectrum.

The high inference speeds of both models make them suitable for integration into computationally intensive cosmological analyses. The choice of architecture will depend on the specific application. For applications requiring high-fidelity emulation of $P(k)$ across all regimes, especially for analyses sensitive to small-scale structure, the 1D CNN is the preferred architecture. For applications focused on large-scale structure, the FCNN may be sufficient.

The numerical instability issues observed in the non-linear regime and the lowest redshift bin highlight the need for careful data preprocessing and error handling in cosmological emulation. Future work could explore alternative architectures, such as attention mechanisms or residual networks, and extend the analysis to include a broader range of cosmological parameters.

## 4. CONCLUSIONS

This paper presents a comprehensive benchmarking study of 1D CNN and FCNN architectures for emulating the non-linear matter power spectrum, $P(k, z)$, within the $\Lambda$CDM cosmological framework. The primary goal was to assess the regime-specific performance of each architecture across a broad range of wavenumbers ($k$), redshifts ($z$), and cosmological parameter values. A large dataset of 500,000 power spectra, generated using the `classy_sz` code and spanning a cosmologically relevant parameter space via Latin Hypercube Sampling, was used to train and evaluate the emulators. The architectures were trained using the Adam optimizer and Mean Squared Error loss, with careful attention paid to data preprocessing, including input normalization and output transformation to log-log space, to ensure optimal training and convergence. Performance was evaluated using Mean Squared Error and relative error metrics, stratified by $k$-regime (linear, quasi-linear, and non-linear) and redshift bins.

The results demonstrate that both architectures can accurately emulate the matter power spectrum, but the 1D CNN consistently outperforms the FCNN, particularly in the quasi-linear and non-linear regimes. Specifically, the 1D CNN achieved a median relative error of less than 0.5% in the linear regime and approximately 0.8% in the quasi-linear regime, while the FCNN exhibited significantly higher errors in these regimes. Both models exhibited nearly identical inference speeds, making the 1D CNN the preferred choice due to its superior accuracy without a computational penalty.

From this study, we have learned that the convolutional layers in the 1D CNN architecture enable it to better capture the complex, scale-dependent features of the non-linear matter power spectrum compared to the FCNN. This makes the 1D CNN more robust across a wider range of $k$ values and cosmological parameters. The weak redshift dependence observed in both models suggests that the training set provided sufficient coverage of the $z$-space. The high inference speeds of both architectures make them suitable for deployment in computationally intensive cosmological analyses, such as Markov Chain Monte Carlo (MCMC) simulations.

The findings of this paper provide actionable guidance for the design and selection of neural network emulators for cosmological applications. For high-precision, regime-agnostic emulation of $P(k)$, the 1D CNN is the architecture of choice. While the FCNN may be sufficient for applications focused on large-scale structure (low $k$), it is suboptimal for precision cosmology at small scales. Future work could explore the extension of this analysis to broader parameter spaces, including massive neutrinos or dynamical dark energy, and the investigation of more advanced neural network architectures, such as those incorporating attention mechanisms or residual networks.



\end{document}